\documentclass[10pt,twocolumn,letterpaper]{article}

\usepackage[cameraready]{cvpr} 

\usepackage{graphicx}
\usepackage{amsmath}
\usepackage{booktabs}
\usepackage{amssymb}
\usepackage{marvosym}
\usepackage{stfloats}
\usepackage{multirow}

\usepackage{colortbl}
\usepackage{pifont}
\usepackage{tikz}
\usepackage{xcolor}
\usepackage{pgfplots}
\usepackage{tcolorbox}
\usepackage{algorithm}
\usepackage{algorithmic}

\usepackage{amsmath}
\usepackage{xcolor}
\definecolor{citecolor}{HTML}{0071bc}
\usepackage[pagebackref=false,breaklinks=true,letterpaper=true,colorlinks,citecolor=citecolor,bookmarks=false]{hyperref}
\usepackage{enumitem}
\usepackage[capitalize]{cleveref}
\crefname{section}{Sec.}{Secs.}
\Crefname{section}{Section}{Sections}
\Crefname{table}{Table}{Tables}
\crefname{table}{Tab.}{Tabs.}

\newcommand{\our}{\textup{VLog}}
\newcommand{\cmark}{\ding{51}}%
\newcommand{\xmark}{\ding{55}}%

\def\cvprPaperID{477} 
\def\confName{CVPR}
\def\confYear{2024}

\definecolor{citecolor}{HTML}{0071bc}
\definecolor{fuchsia}{rgb}{0.6, 0.33, 0.73}
\usepackage[pagebackref=false,breaklinks=true,letterpaper=true,colorlinks,citecolor=citecolor,bookmarks=false]{hyperref}

\definecolor{NiceBlue}{rgb}{0.11764705882352941, 0.5647058823529412, 1.0}
\definecolor{Gray}{gray}{0.9}
\hypersetup{
    colorlinks=citecolor,
    linkcolor=citecolor,      
    urlcolor=citecolor         
}
\begin{document}

\title{\our: Video-Language Models by Generative Retrieval of\\Narration Vocabulary}

\author{Kevin Qinghong Lin,
Mike Zheng Shou\textsuperscript{\Letter}
\\\\
Show Lab, National University of Singapore
}
\maketitle
\newcommand\blfootnote[1]{%
  \begingroup
  \renewcommand\thefootnote{}\footnote{#1}%
  \addtocounter{footnote}{-1}%
  \endgroup
}

\begin{abstract}
Human daily activities can be concisely narrated as sequences of routine events (e.g., turning off an alarm) in video streams, forming an event vocabulary. 
Motivated by this, we introduce \our, a novel video understanding framework that define video narrations as vocabulary, going beyond the typical subword vocabularies in existing generative video-language models. 
Built on the lightweight language model GPT-2, 
\our\ feature three key innovations:
{\textbf{(i) A generative retrieval model}, marrying language model's complex reasoning capabilities with contrastive retrieval's flexible upgrading over narration vocabulary}.
\textbf{(ii) A hierarchical vocabulary} derived from large-scale video narrations using our narration pair encoding algorithm, enabling efficient indexing of specific events (\eg, cutting a tomato) by identifying broader scenarios (\eg, kitchen) with expressive postfixes (\eg, by the left hand).
\textbf{(iii) A vocabulary update strategy} leveraging generative models to extend the vocabulary for novel events encountered during inference.
To validate our approach, we introduce VidCap-Eval, a development set requiring concise narrations with reasoning relationships (\eg, before and after).
Experiments on EgoSchema, COIN, and HiREST further demonstrate the effectiveness of \our, highlighting its ability to generate concise, contextually accurate, and efficient narrations, offering a novel perspective on video understanding.
Codes are released at \url{https://github.com/showlab/VLog}.
\end{abstract}

\blfootnote{\Letter: Corresponding Author.}

\vspace{-3.5em.}
\section{Introduction}
\begin{figure}[!t]
\centering
\includegraphics[width=\linewidth]{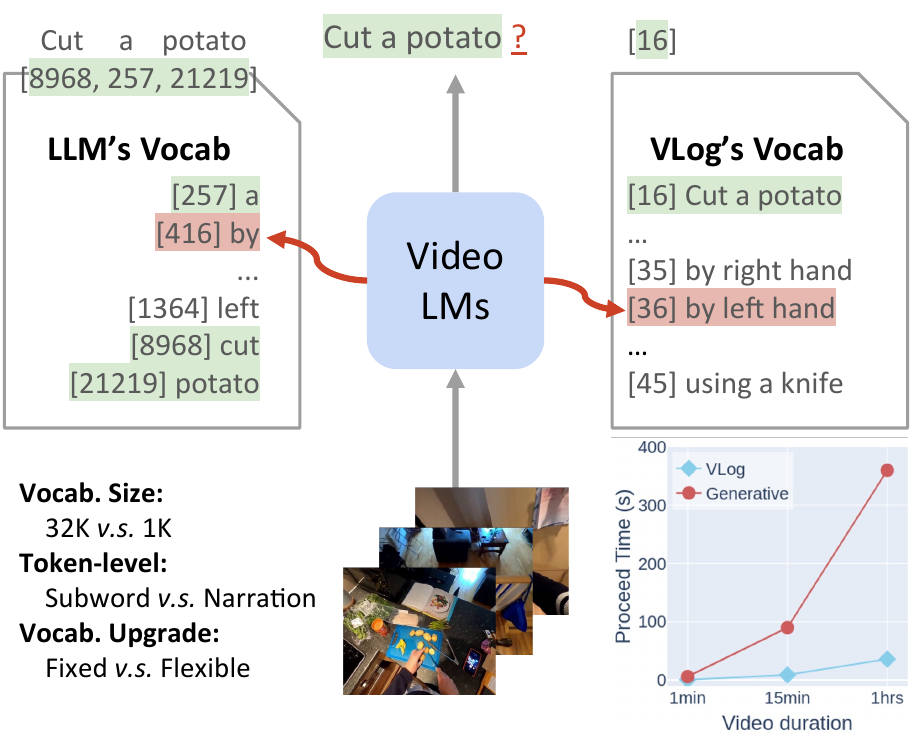}
\vspace{-2em.}
\caption{\textbf{Key Idea of \our.} 
In contrast to existing video-language models that rely on token-by-token generation on language's subword vocabulary, \our~introduces a novel \textbf{generative retrieval method based on a narration vocabulary}, achieving a significant speedup (10$\times$) when processing long videos.
}
\vspace{-2.5em}
\label{fig:teaser}
\end{figure}

\textit{``Life is a succession  of moments."} -- Corita Kent

Recent advances in Large Language Models (LLMs)\cite{llama3,llama31} have inspired several research into transferring textual knowledge to multi-modal domains~\cite{llava,minigpt}, leading to the development of Video Large Language Models (VideoLLM)~\cite{flamingo,videochatgpt}. These models mainly leverage foundational elements of LLMs, such as subword vocabularies~\cite{bpe} and transformers with pretrained weights, then adapt them through multi-modal instruction tuning~\cite{llava,videochat}. As a result, these models can generate video-conditioned textual outputs via next-token prediction, as illustrated on the left side of Fig.~\ref{fig:teaser}.

Despite these advancements, original LLM designs are not inherently suited for video understanding. For example, LLM subword vocabularies are typically large (\eg, LLama-3’s 128K vocabulary size~\cite{llama3}) to capture broad linguistic information, but incomplete subwords (\eg, ‘happ’) often lack visual interpretability. Moreover, token-wise generation during inference introduces a bottleneck, limiting the model’s ability to process video in real time.

In practical applications, models are not always required to provide exhaustive details. Instead, we often prefer a video model that delivers concise, contextual responses in real time—such as an AR glasses assistant customized for personal needs \ie, prioritize \textbf{task-specific efficiency} over generalist models.
This raises the question: How can we tailor a video model to meet our specific requirements? 
To address this, we draw inspiration from how humans naturally organize experiences. When reflecting on our day, we often recall it as a series of narrative events—such as washing dishes or reading a book—that form the ‘narration vocabulary’ of our daily lives.

Motivated by this, we propose \our, a novel efficient video understanding framework.
Unlike generative VideoLLM that rely on subword vocabularies, \our~represents video narrations as minimal token units and builds the vocabulary. 
As shown on the right side of Fig.\ref{fig:teaser}, this narration vocabulary results significantly reduces decoding times.

To leverage the narration vocabulary in LLMs, it need to train the embedding layers to accommodate newly added vocabulary units.
In contrast, CLIP~\cite{clip}-based retrieval models can directly extract embeddings for novel vocabulary without additional training, enabling efficient similarity matching.
However, retrieval models often lack the reasoning capabilities of generative models, failing to handle advanced queries like “What’s the next action?” with a reference video.
To address this issue, \our\ introduces a novel \textbf{generative retrieval} architecture that repurposes the language model's reasoning capabilities with a retrieval token, embedding both visual and query information for reasoning-oriented retrieval.

To construct the narration vocabulary, we develop a Narrative Pair Encoding method applied to exisiting video narration datasets~\cite{ego4d}, generating prefix sets (\eg, Cut a potato) and postfix sets (\eg, by the left hand). Additionally, to enable efficient indexing across a large vocabulary, we organize the vocabulary hierarchically instead of Brute-force search. This design allows rapid indexing of prefix vocabulary subset (\eg, ‘Cut a tomato’) by first identifying scenarios (\eg,‘kitchen’) and then refining the search with the postfix, as illustrated in Fig.\ref{fig:teaser}.

Recognizing that the initial vocabulary may not cover novel, out-of-vocabulary events, we devised a vocabulary upgrade strategy using an agentic workflow. When low similarity scores are detected for vocabulary entries, they are treated as novel events and processed by generative models like LMM~\cite{llavaov} to generate scene descriptions. LLM~\cite{qwen2.5} is then prompted to extend the relevant vocabulary entries.

Overall, our contributions are three folds:
\begin{itemize}[itemsep=0.5em, parsep=0em, topsep=0em]
\item {We introduce the first narration vocabulary for VideoLLMs, which is constructed by our Narration Pair Encoding method and a hierarchical strategy to enable efficient indexing, resulting xxx speedup;
}
\item We introduce a novel generative retrieval strategy, which supports flexible vocabulary upgrading without training embedding for new vocabulary. 
\item We demonstrate the effectiveness and efficiency of \our\ on the new development set Vidcab-Eval and public datasets, including COIN, EgoSchema, and HiREST, highlighting its strengths in efficient and accurate video understanding.
\end{itemize}
\section{Related Works}
\subsection{Video Captioning}
Video captioning provides a natural way to interpret and describe video content, with a key focus on understanding human activity~\cite{handscale, egovocab, visor}. This includes various downstream tasks, such as dense captioning~\cite{vidchapters} and step captioning~\cite{coin}. Some video captioning datasets emphasize long, descriptive paragraphs that aim to capture every detail~\cite{activitynet}.
While in real-world applications, brief, contextual sentences are often sufficient to convey what happens in a video—essentially, \textit{video narration}. First highlighted in~\cite{ego4d} as an efficient alternative for documenting untrimmed video streams, video narration functions like a “minute”, capturing key events or changes with sparse yet informative content over time.
With a large and diverse set of video narrations, as in ~\cite{ego4d,egoexo4d}, we can reasonably assume that most common daily activities, such as “turning off an alarm upon waking”, can be reused across contexts, much like human experience. 
In our work, we treat these narrations as a vocabulary of human behavior, using them to interpret new, incoming video streams.

\subsection{Video-Language Models}
\noindent\textbf{Retrieval Models.}
Early vision-language models~\cite{clip,ht100m,webvid,egovlp,internvideo} primarily relied on alignment approaches, leveraging contrastive objectives for scalable performance. 
These methods excelled in retrieval and classification tasks within pre-defined labeled spaces, offering high efficiency by {narration-level} dot-product similarities calculation~\cite{glip,helpinghands,egovlpv2,univtg}. 
But this fashion lacks the ability to model complex reasoning relationships between media inputs;
for example, they fail to retrieve a video clip based on a query like, “What occurred after?” which conditioned on a reference video clip.

\noindent\textbf{Generative Models.}
Recently, focus has shifted to generative LMMs, with several works~\cite{flamingo,blip2,llava,minigpt,instructblip,videochat,showo} developing large multimodal models by projecting visual inputs into textual embeddings and aligning them with LLMs through visual instruction tuning. 
However, such straightforward adaptations are often unsuitable for video understanding, as large {subword-level} vocabularies in LLMs, while broadly inclusive, lack visual interpretability. Moreover, slow decoding during testing hinders real-time applications.
Therefore, recent studies have focused on improving the efficiency of VideoLLM for (long) video modeling~\cite{moviechat,llamavid,videollmol,videollmmod,longctxv,longvu}.

\noindent\textbf{Generative Retrieval.}
By examining the strengths and limitations of retrieval and generative models, a natural approach is to unify them, leveraging reasoning alongside efficient retrieval. In the realm of LLMs, efforts~\cite{e5,tart,instructor,nvembed} have been made to improve text embeddings that capture complex relationships, showing promise for practical applications. However, these advancements are less explored in multimodal~\cite{fromage,mmembed,vlm2vec}, particularly in video understanding.

In this work, we pursue this generative retrieval direction by introducing a novel method that marrying lightweight GPT-2’s reasoning~\cite{gpt2} with the contrastive vision-text model SigLIP~\cite{siglip}. Moreover, we departing from traditional subword vocabularies, \our\ redefines the narration vocabulary, incorporating retrieval-based search for {flexible} vocabulary upgrading and interpretable reasoning.
\begin{figure*}[!t]
\centering
\includegraphics[width=1.0\linewidth]{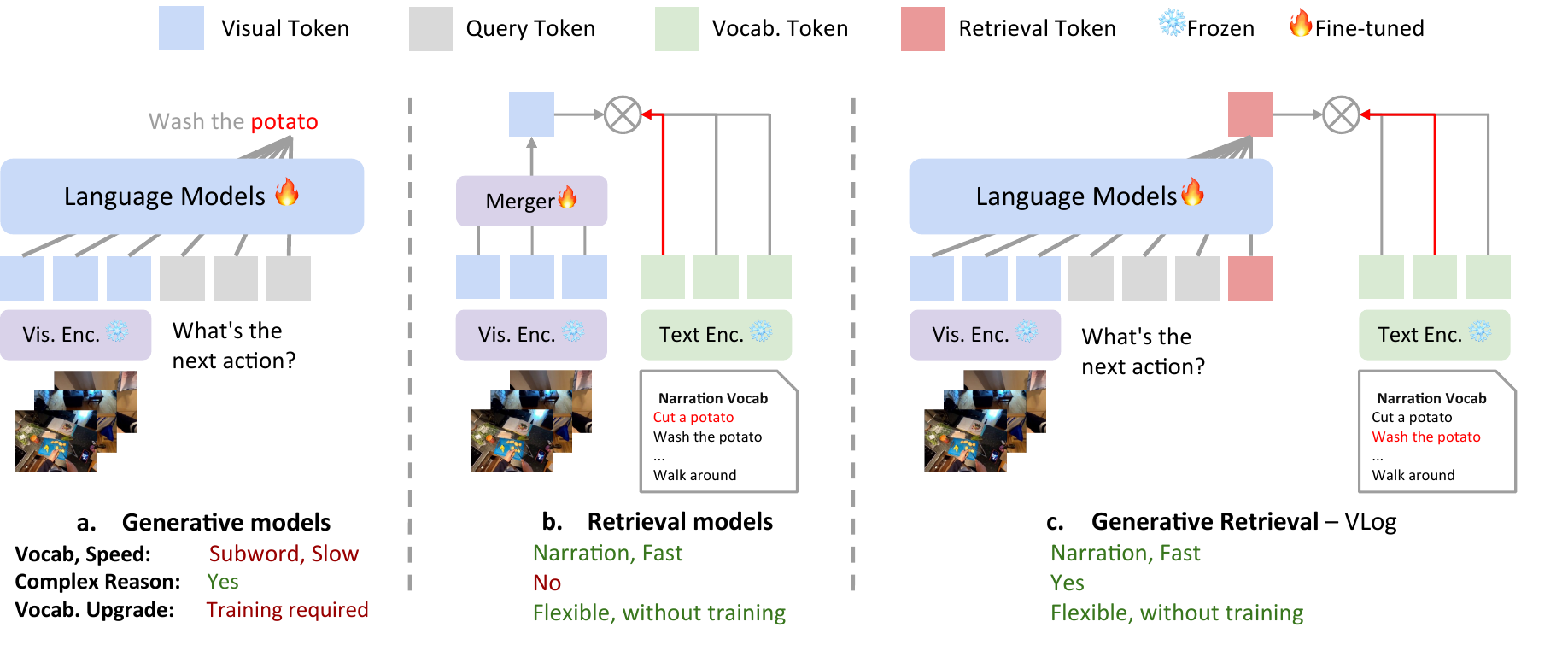}
\caption{
\textbf{Comparison between different Video-Language model architectures:} 
\textbf{(a) Generative Models:} These model with complex reasoning but are slow, generating {subword} tokens one by one, Vocabulary upgrading requires retraining.
\textbf{(b) Retrieval Models:} These enable {fast narration vocabulary via similarity matching} and support vocabulary upgrading without training, but lack reasoning, useful only for simple alignment tasks.
\textbf{(c) Generative Retrieval (VLog):} This approach combines fast narration-vocabulary {and flexible} vocabulary upgrading with complex reasoning by using a retrieval token, merging the advantages of both methods.
}
\label{fig:model}
\vspace{-2em.}
\end{figure*}
\section{\our}
\noindent\textbf{Task definitions:}
Given a video \(\mathcal{V} = \{v_i\}\), where each \(v_i\) represents a frame, and a conditional query \(\mathcal{Q}\), our objective is to generate an accurate narration \(\mathcal{X}\) of the video. To ensure consistency across all variants, we avoid specialized frame sampling strategies, focusing instead on the distinct strengths of generative and retrieval modeling.

\subsection{Architecture}
\label{sec:arch}
\noindent\textbf{a. Generative Models.}
In most generative approaches, \ie, large multimodal models, a subword vocabulary \(\mathcal{O} = \{o_i\}\) from language models is typically used, where \(L = |\mathcal{O}|\) denotes the vocabulary size (each \(o_i\) represents a subword, such as 
‘the’).
Given visual inputs and a query encoded into a token sequence, the language model estimates the probability of the next token \({x}_n\) being the \(i\)-th token $o_i$ over the vocabulary $\mathcal{O}$, conditioned on the visual inputs $\mathcal{V}$, query $\mathcal{Q}$, and previously generated tokens ${x}_{<n}$:
\begin{equation}
\Pr\left({x}_n = o_i \mid {x}_{<n}, \mathcal{V}, \mathcal{Q}\right) \quad \text{for} \quad o_i \in \mathcal{O},
\end{equation}
where $\Pr$ is parameterized by language model weights using cross-entropy.
Normally, the next token is predicted by maximum likelihood:
\begin{equation}
\tilde{{x}}_n = \arg\max_{i} \Pr({x}_n = o_i \mid {x}_{<n}, \mathcal{V}, \mathcal{Q}).
\end{equation}

This autoregressive process enables the model to capture complex dependencies between the visual inputs and the query, as demonstrated in Fig.\ref{fig:model}(a).
However, for dense narrations over a long video, this approach incurs high inference costs due to token-by-token decoding~\cite{llovi, videoagentlong, videoagentmem}. 
Despite the efficiency gains brought by adopting the narration-level vocabulary, adding new vocabulary units to the embedding layer of LLMs typically requires training, making the approach less flexible.

\noindent\textbf{b. Retrieval with Vocabularies.}
Inspired by how humans retrieve past experiences to interpret new coming events, we propose reframing the token generation process as a retrieval task with a predefined narration set \(\tilde{\mathcal{O}}=\{\text{cut a potato}, \cdots, \text{walk round}\}\), serving as a behavior vocabulary as depicted in Fig.\ref{fig:model}(b).
We employ a vision-text contrastive model, SigLIP~\cite{siglip} as it similarity matching can support flexible vocabulary upgrading. The model maps video frames and vocabulary tokens into a shared embedding space, yielding vocabulary embeddings \(\tilde{\mathbf{o}_i} \leftarrow \text{SigLIP}(o_i)\) and frame embeddings \(\mathbf{v}_j \leftarrow \text{SigLIP}(v_j)\).
To model temporal relationships among frames, 
we use an additional module—
a 2-layers transformer layers-
to produce a compact clip representation \(\tilde{\mathbf{v}}\leftarrow\{\mathbf{v}_j\}\) from the sequence of frame embeddings. 

To sample prediction by leveraging the narration vocabulary, we define the probability between a narration vocabulary \(\tilde{o_i} \in \tilde{\mathcal{O}}\) and the video clip \(\tilde{\mathbf{v}}\) based on their cosine similarity over their projected embeddings:
\begin{equation}
\Pr\left(\mathcal{X} = \tilde{o_i} \mid \mathcal{V}\right)=\tilde{\mathbf{v}}^T\tilde{\mathbf{o}_i} \quad \text{for} \quad \tilde{o_i} \in \mathcal{O},
\label{dotproduct}
\end{equation}
the prediction is then determined by
$\tilde{\mathcal{X}} = \arg\max_{o_i} \left(\tilde{\mathbf{v}}^T\tilde{\mathbf{o}_i}\right)$

This approach offers two benefits: 
first, it enables narration-level retrieval via dot product similarity calculation (\ie, Eq.~\ref{dotproduct});
second, the vocabulary is independent of LLM embedding weights, making it practical for upgrades when needed.
However, despite leveraging the strengths of contrastive models, this method struggle to capture the complex relationships in queries $\mathcal{Q}$, posing challenges for retrieving with a queries \eg, ‘What happened next.’

\noindent\textbf{c. Generative Retrieval (Ours).}
{To leverage the reasoning modeling of generative models while harnessing the efficiency of narration-vocabulary and the flexiblity of vocabulary upgrading, we propose a novel generative retrieval model.}
As shown in Fig.\ref{fig:model}(c), we introduce a \textit{retrieval token} with embedding \(\mathbf{t}\) to bridge the generative language model and retrieval model. The retrieval token is positioned as the \textit{last} input in the language model sequence, allowing it to attend to both the front visual and query inputs. After passing through the language model, the output embedding \(\tilde{\mathbf{t}}\) is assumed to encode both visual and query information, enabling retrieval while preserving reasoning ability:
\begin{equation}
\Pr\left(\mathcal{X} = \tilde{o_i} \mid \mathcal{V}, \mathcal{Q}\right)=\tilde{\mathbf{t}}^T\tilde{\mathbf{o}_i} \quad \text{for} \quad \tilde{o_i} \in \tilde{\mathcal{O}},
\end{equation}
the prediction is then yield by
$\tilde{\mathcal{X}} = \arg\tilde{\max_{o_i}} \left(\tilde{\mathbf{t}}^T\tilde{\mathbf{o}_i}\right)$.

Moreover, our model has the following considerations:
\begin{enumerate}
\item \textbf{Retrieval tokens initialization:} The retrieval token, appearing at the end of the sequence, can be initialized as a learnable token, an EOS token, or using mean-pooled visual inputs to encode overall visual information—examined in our experiments.
\item \textbf{Asymmetric structure:} Unlike visual and query inputs, vocabulary embeddings are not projected by the language model and remain fixed after initial computation, reducing forward computation costs with large vocabularies.
\end{enumerate}
In this way, our architecture effectively addresses both reasoning and efficiency issues.

\noindent\textbf{Training objectives.} 
Our generative retrieval model is trained with a standard contrastive objective:
\begin{equation}
\mathcal{L}=\frac{1}{|\mathcal{B}|}\sum_{i\in \mathcal{B}} \log \frac{\exp(\tilde{\mathbf{t}}_i^T\tilde{\mathbf{o}_i} /\tau)}{\sum_{j\in \mathcal{B}} \exp( \tilde{\mathbf{t}}_i^T\tilde{\mathbf{o}_j} /\tau)},
\label{nce}
\end{equation}
where \(\mathcal{B}\) is the batch size and \(\tau = 0.05\) is the temperature.
This objective updates the language model weights to encourage retrieval tokens $\tilde{\mathbf{t}}$ derived from visual and textual inputs to align with the target vocabulary $\tilde{\mathbf{o}_i}$.

\begin{figure}[!t]
	\centering
	\includegraphics[width=1.0\linewidth]{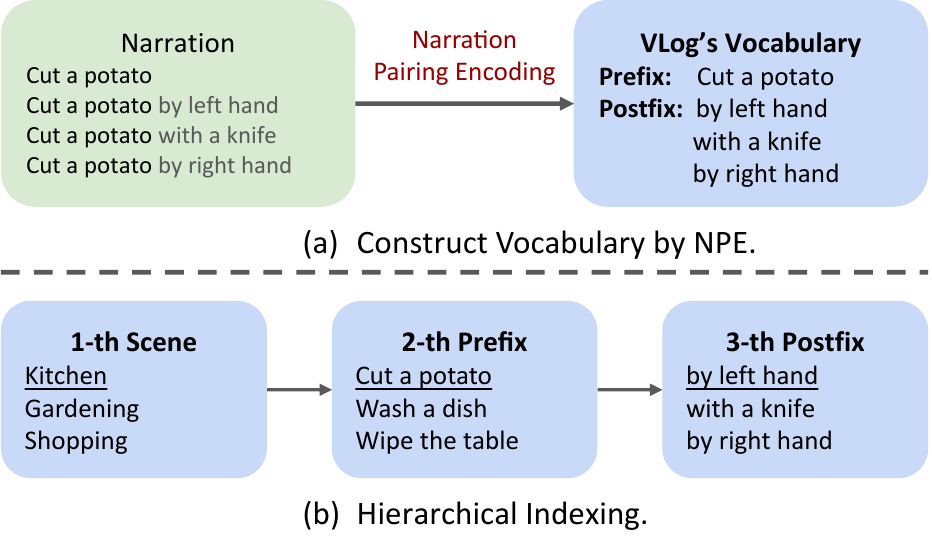}
    \vspace{-2em.}
    \caption{
\textbf{Illustration of our Vocabulary Construction and Indexing.}
\textbf{Upper side:} Given the narrations, we process them using our NPE method, breaking down each narration into prefix and postfix parts.
\textbf{Lower side:} For efficient indexing, we organize the vocabulary hierarchically, where first-level scenes help navigate subsets of prefix narrations. Next, we append the prefix and continue retrieving the postfix.
}
\label{fig:vocab}
\end{figure}

\subsection{Vocabulary Construction}\label{sec:vocab:source}
The narration vocabulary is key to our model. We source it from existing video datasets~\cite{ego4d}, which contain extensive narrations across various domains. We then clean these narrations and remove duplicates (see details in Supp).

\noindent\textbf{Narration Pair Encoding.}
We observe that many narrations are often subjective and inconsistently formatted, as shown in Fig.~\ref{fig:vocab} upper side, they share common prefixes (\eg~“cut a potato”) but differ in their postfixes, which add context. 
In natural language processing, Byte Pair Encoding (BPE)~\cite{bpe} addresses this issue by tokenizing text corpora into subword units.
However, in our setting, we aim to build a narration-level vocabulary where BPE is not directly applicable. 
To address this, we introduce Narration Pair Encoding (NPE):
We treat each narration as a potential prefix and search for longer narrations that start with it, collecting their postfixes—the additional words beyond the prefix. 
This approach yields two sets—\textit{a prefix set} of narrations with non-empty postfixes and a shared \textit{postfix set}. We detailed NPE algorithm in our Supplement.

To enable model for prefix and postfix retrieval, we first use the retrieval token to retrieve the prefix, concatenate it with the visual and query sequence, then use the retrieval token to retrieve a postfix from the postfix vocabulary.

\noindent\textbf{Hierarchical Indexing.}\label{sec:vocab:index}
After completing the NPE process, we obtain both prefix and postfix narration sets. However, the large scale of narrations (\eg, millions) poses a challenge for efficient retrieval, making brute-force search impractical. Considering that human activity recordings in videos often align with specific scenarios, such as “cut a potato” occurring in a “kitchen” scene, we develop a hierarchical indexing strategy by associate narrations to its videos belonged scenario. 

The \textit{full retrieval chain} is displayed in the lower part of Fig.\ref{fig:vocab}. Given a video, the model first identifies the video scenario and then retrieves the associated prefix narrations subset. This is efficient by reducing the search space. Then we continue retrieval the postfix from the postfix vocabulary. This helps to improve the narration expression.

\subsection{Instruction-Tuning Data}
\label{sec:trainingdata}
Once we developed such a generative retrieval model, the next issue we faced was the lack of training data with complex reasoning relationships, as most video-text retrieval data are paired solely for alignment. Fortunately, untrimmed video streams offer a natural solution by inherently modeling temporal relationships between narrations.

\begin{figure}[!h]
	\centering
	\includegraphics[width=1\linewidth]{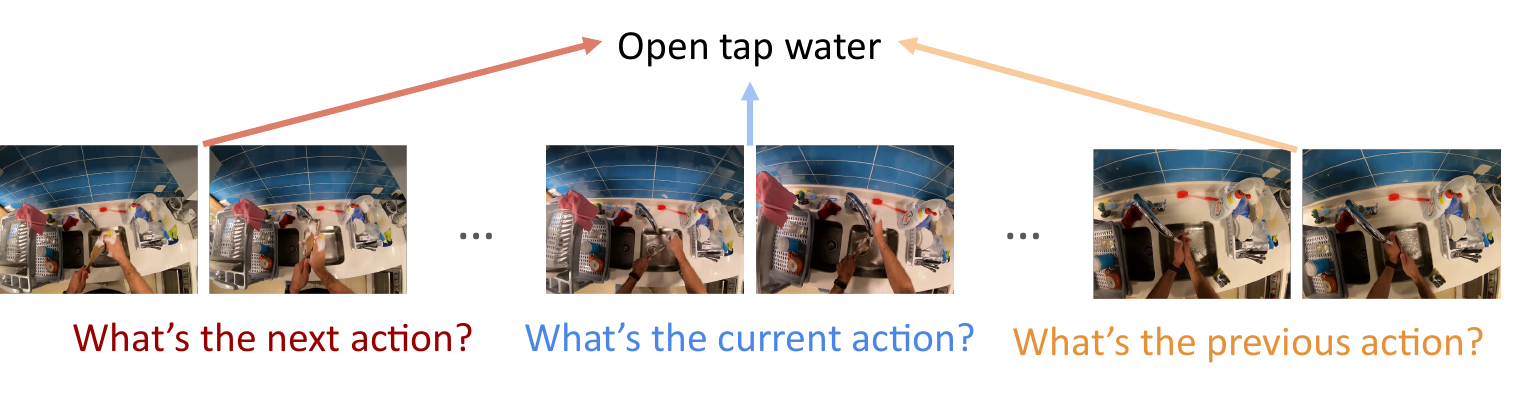}
\caption{
{\textbf{Create video-text pairs that requires complex reasoning} from untrimmed videos based on their temporal relationship.}
}
\label{fig:trainingdata}
\end{figure}

As illustrated in Fig.~\ref{fig:trainingdata}, we develop three types of training data representing ‘before’, ‘next’, and ‘current’ relationships. 
For ‘next’ as an example, given a video clip \(\mathcal{V}_{i}\) with its reference narration \(\mathcal{X}_i\), we trim a preceding segment \(\mathcal{V}_{<i}\) and append a prompt such as “What's the next action?”
Based on these temporal relationship, we create an 200K instruction-tuning training data from ~\cite{ego4d}, namely \textbf{Vidcab-Train} (\ie, \textbf{Vid}eo vo\textbf{cab}ulary). Moreover, as we lack of such retrieval evaluation consider complex reasoning, we apply the same strategy but create a generative retrieval development set with 4.6K samples named as \textbf{Vidcab-Eval}.
Notably, we carefully curate the instruction training data and development set using our NPE method. For instance, in Vidcab-Eval, the model must accurately retrieve both the prefix and postfix to achieve a higher score.
We detail how to construct them in Supp.

Moreover, we incorporate scenario information from videos, prompting the model to encode the entire sequence of a video with queries like “What's the overall activity in this video?” and answers such as “Cooking.” 
This organization supports hierarchical vocabulary indexing, which we will discuss in the next section.

\subsection{Vocabulary Upgrading}\label{sec:vocab:upgrade}
Despite the diversity and scale of initial vocabulary, novel \textbf{Out-of-Vocabulary (OOV)} events may still occur, requiring: (i) detection of novel events, and (ii) expansion of the vocabulary with new entries.

\noindent\textbf{Detect Novel Events.}
Our model uses the dot product in Eq.~\ref{dotproduct} as a relevance metric between the query and candidate vocabulary, akin to logits as a confidence measure in LLMs. Empirically, we set a threshold of 0.4: if the top-1 vocabulary match falls below this threshold for a given visual input, we classify the event as OOV.

\begin{figure}[!t]
	\centering
	\includegraphics[width=1\linewidth]{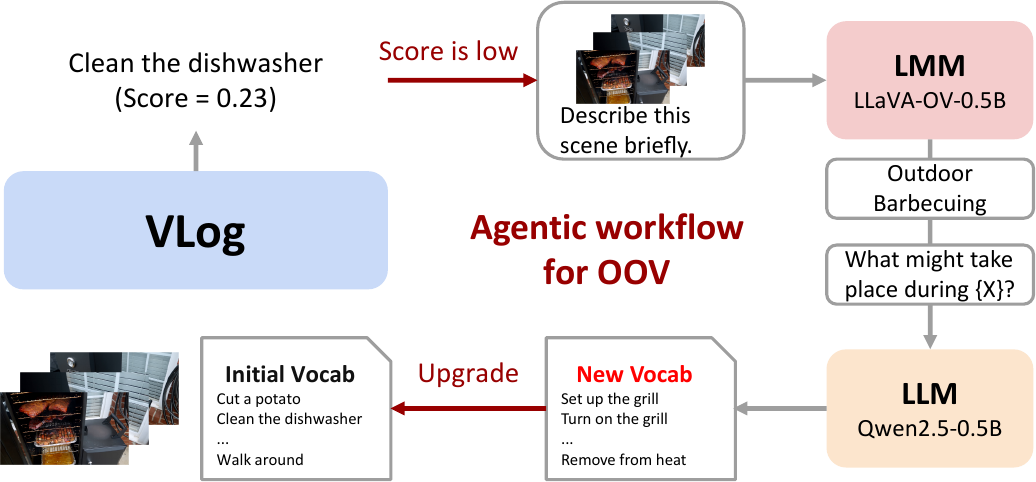}
	\caption{
\textbf{Agentic workflow of VLog Vocabulary upgrade}. 
When a low retrieval score is detected, the visual inputs are sent to the LMM (LLaVA-OV-0.5B~\cite{llavaov}) to generate scene descriptions. These descriptions are then processed by the LLM (Qwen2.5-0.5B~\cite{qwen2.5}), which expands and updates the existing vocabulary.
}
\label{fig:update}
\end{figure}

\noindent\textbf{Seeking Help from Generative Models.}
To handle out-of-vocabulary narrations, we leverage both LMMs and LLMs, which offer complementary strengths: LMMs generate brief, visually grounded captions but may hallucinate on fine-grained actions, while LLMs, with broad knowledge and strong instruction-following, generate diverse candidate narrations. 
We integrate these models in an \textit{agentic} workflow.
As illustrated in Fig.\ref{fig:update}, 
where LMMs first provide concise visual scene descriptions, and LLMs then use this context to infer potential events that might occur in the scene, which are parsed as new vocabulary narrations.
In practice, we use two models comparable in size to our GPT-2: the LMM -- LLaVA-OV-0.5B~\cite{llavaov}, and the LLM -- Qwen2.5-0.5B~\cite{qwen2.5} guided by an in-context template. 

This workflow differs from existing Retrieval-Augmented Generation~\cite{ragllm}, serving instead as a new {Generative-Augmented Retrieval} approach, where generative models actively expand the vocabulary for improved retrieval accuracy.
\section{Experiments}
\noindent In this section, we structure our experiments to answer the following questions:

\noindent \textbf{Q1:} What key advantages does \our~offer?

\noindent \textbf{Q2:} How to adapt \our~to different tasks~(\eg, Reasoning QA beyond retrieval)?

\noindent \textbf{Q3:} How effective is the vocabulary upgrading strategy (address out-of-vocabulary problem)?

\noindent \textbf{Q4:} What are the key design choices in \our~(\eg, NPE and Vocabulary indexing)? 

\subsection{Datasets and Settings}
\noindent
\textbf{Vidcab-Eval} is constructed by Sec.~\ref{sec:trainingdata}, we selected Ego4D~\cite{ego4d} to build our initial vocabulary and training data due to its large scale and diversity, with millions of manually curated narrations. 
We ensure no overlap with downstream tasks such as EgoSchema.
To evaluate our proposed generative retrieval setting, we evaluate on Vidcab-Eval development set.

\noindent
\textbf{COIN} is an instructional video dataset comprising 11,827 videos across 180 tasks in 12 domains related to daily life. 
We evaluate our model on three common COIN benchmarks: step recognition, step forecasting, and task summarization.
We use this benchmark to study \our's ability for fine-grained activity recognition.

\noindent
\textbf{EgoSchema} is a long-range video question-answering benchmark with 5K multiple-choice pairs across 250 hours of video, covering a wide range of human activities. Unlike prior benchmarks, correctly answering a question here requires at least 100 seconds of video viewing—well beyond existing standards. We use this benchmark to study \our's high-level reasoning abilities.

\noindent
\textbf{HiREST} is a new benchmark for hierarchical procedural information. 
It includes videos from novel domains that not appeared in Ego4D and COIN, with numerous and high-quality step captions by human annotators.  Thus, we use its step-captioning task to study the effectiveness of \our's vocabulary upgrading strategy.

\begin{table*}[!t]
\centering
\resizebox{0.9\textwidth}{!}{
\begin{tabular}{llllccccccl}
    \toprule
    & \multicolumn{3}{c}{\textbf{Models}} & \multicolumn{1}{c}{\textbf{FT?}} & \multicolumn{2}{c}{\textbf{Navie Retrieval}} & \multicolumn{2}{c}{\textbf{Casual Retrieval}} & \textbf{Decode time} \\ 
    & Visual Enc. & Post process & Vocabulary & & CIDEr & R@1 & CIDEr & R@1 & sec\\
    \midrule
    Generative & SigLIP-L & GPT2-M & GPT2-32K & \cmark & 64.8 & 7.9 & 53.7 & 3.1 & 0.362 \\
    \textcolor{gray}{Retrieval} & \textcolor{gray}{SigLIP-L} & \textcolor{gray}{MeanPool} & Eval-4.6K & \textcolor{gray}{\xmark} & \textcolor{gray}{63.6} & \textcolor{gray}{4.6} & \textcolor{gray}{N/A} & \textcolor{gray}{N/A} & {0.001} \\
    Retrieval & SigLIP-L & Adapter & Eval-4.6K & \cmark & 95.8 & 11.8 & 48.9 & 2.1 & \textbf{0.016} \\
    \hline
    \our & SigLIP-L & GPT2-M & Eval-4.6K & \cmark & \textbf{96.9} & \textbf{12.4} & \textbf{87.3} & \textbf{5.0} & \textbf{0.018 (20$\times$)} \\
    \our-prefix & SigLIP-L & GPT2-M & Ego4D-0.8M & \cmark & 91.3 & 10.9 & 83.9 & 3.7 &  \underline{0.035 (10$\times$)} \\
    \our-prefix\&postfix & SigLIP-L & GPT2-M & Ego4D-0.8M & \cmark & \underline{94.9} & \underline{11.9} & \underline{86.9} & \underline{4.8} & 0.053 (6$\times$) \\
    \bottomrule
\end{tabular}
}
\caption{\textbf{Key ablation studies on Vidcab-Eval for naive and casual retrieval.} Under same conditions, \our~ provides accurate narration with significant speed improvements.
}
\vspace{-1em.}
\label{fig:ablation:key}
\end{table*}

\noindent\textbf{Implementation Details.} \our~builds on GPT2-medium\cite{gpt2} with SigLIP~\cite{siglip}, extracting video clips at 2 frames per second.
For long videos, we uniformly sample long video into multiple fix length clips (1s) and process them in a streaming fashion.
We pre-extract the visual features and store the textual embedding of narration vocabulary, which storage occupies 3.9 GB in total.
During training, we fine-tune the models fully.

\subsection{Key Advantages by VLog}
In Tab.~\ref{fig:ablation:key}, we evaluate key variants in \our-Eval, which aims to produce accurate narrations based on both visual and textual queries. 
This includes two tracks: a naïve track (without query, \ie~normal video-text alignment) and a causal track (with query, \eg, before / after / current). To enable both generative and retrieval models to report scores, we introduce two metrics to assess narration quality: CIDEr (common used by generative models) and Recall@1 (common used by retrieval models) alongside generation speed per clip for a comprehensive evaluation.

\noindent\textbf{Baselines:} 
To ensure a fair comparison, all variants use the same visual encoder, and both the generative and \our~models are based on the same GPT2 model trained on the same data.
For the retrieval baseline, we provide both zero-shot and fine-tuned versions, using \our-Eval-4.6K as the restricted vocabulary, ensuring that the correct term is within the vocabulary. 
To adapt the retrieval model for casual retrieval, we pool the visual and textual query inputs to obtain a unified embedding. 
For \our, we provide variants using either Eval-4.6K or our full constructed vocabulary, with or without postfix retrieval.
We have the following observation:

\noindent\textbf{(i)} Overall, the scores in the naïve setting are higher than in the causal setting, indicating that causal retrieval is more challenging than standard alignment, as it requires a fine-grained understanding of temporal relationships.

\noindent\textbf{(ii)} In the naïve setting, \our~achieves comparable CIDEr and Recall@1 scores to retrieval models. In a more challenging causal setting, \our~significantly outperforms both generative and retrieval models, the latter of which pools both visual and query inputs but fails to model relationships effectively. This demonstrates that our generative retrieval approach effectively leverages the language model's causal modeling within a flexible retrieval framework.

\noindent\textbf{(iii)} Using our constructed vocabulary (Ego4D-0.8M, where the ground truth may not be included), the prefix's performance is slightly lower than the baseline (Eval-4.6K), as the full vocabulary may not contain exact matches from the development set. However, when equipped with our postfix, performance improves, demonstrating the high expressiveness of our full narration vocabulary and postfix.

\noindent\textbf{(iv)} In terms of decoding speed, \our~is comparable to the fastest retrieval model, achieving a \textbf{20x} speedup over generative models. Although indexing slows slightly with a large vocabulary (0.8M), it remains significantly faster (10x) than generative models, highlighting its efficiency advantage with the narration vocabulary.

These ablation studies demonstrate the great potential of \our, a video narrator that provide concise narrations with flexible query conditions and fast decoding speed.

\subsection{Adapting VLog to Different Tasks}
Beyond casual retrieval tasks, we next demonstrate how \our~can be adapted to other mainstream tasks: Fine-grained action perception and High-level reasoning QA.

\noindent\textbf{Fine-grained Action Perception: COIN.}
The COIN tasks consist of closed-set action categories, enabling our models to adapt seamlessly by substituting the vocabulary with COIN's predefined category set. Additionally, different setting \ie~steps, tasks, or next actions are matched in \our~by using varied queries as conditions.
We develop a generative GPT2 baseline for comparison, which directly outputs the ground-truth string. All methods are fine-tuned on the COIN dataset, with results provided both with and without Ego4D vocabulary pretraining.

As demonstrated in Tab.~\ref{tab:coin}, \our~outperforms the GPT2 baseline by a significant margin as well as process speed per clip (20$\times$).
With a lightweight model size (124M), \our~achieves performance comparable to the state-of-the-art LMM \cite{videollmol}, built on LLama-2-7B.
Additionally, we demonstrate that generative retrieval pretraining from Ego4D successfully transfers to the COIN dataset.

\begin{table}[!t]
\centering
\scriptsize
\resizebox{0.48\textwidth}{!}{
\begin{tabular}{lclccc}
    \toprule
    \multirow{2}{*}{Method}  & \multirow{2}{*}{PT by?} &  \multirow{2}{*}{Time (s)} & \multicolumn{3}{c}{Top-1 Acc} \\
    & & & Step & Task & Next \\
    \midrule
    ClipBERT~\cite{clipbert}  & COCO+VG & -- & 30.8 & 65.4 & - \\
    TimeSformer~\cite{timesformer} & HT100M & -- &46.5 & 85.3 & 34.0 \\
    Paprika~\cite{paprika}  & HT100M & -- &51.0 & 85.8 & 43.2 \\
    DistantSup~\cite{distantsup}  & HT100M & -- &54.1 & 90.0 & 39.4 \\
    VideoTF~\cite{videotf}  & HT100M &-- & 56.5 & 91.0 & 42.4 \\
    ProcedureVRL~\cite{procedurevrl}  & HT100M & -- &56.9 & 90.8 & 46.8 \\
    VideoTaskGraph~\cite{video_mined_task_graph}  & HT100M &-- & 57.2 & 90.5 & 40.2 \\
    {VideoLLM-online}-7B~\cite{videollmol} & N/A & -- &\textbf{59.8} & {92.1} & {48.1} \\
    \midrule
    GPT2-medium & N/A & 0.21 & 44.6 & 82.4 & 32.1\\
    \our & N/A & \textbf{0.01} & {56.1} & {93.0} & {46.0} \\
    \our & Ego4D & 0.05 & 57.4 & \textbf{94.4} & \textbf{48.4} \\
    \bottomrule
\end{tabular}
}
\caption{\textbf{Activity perception results on COIN benchmarks}: step recognition, task recognition, and next-step forecasting.}
\vspace{-1em.}
\label{tab:coin}
\end{table}

\begin{table}[!b]
\resizebox{0.48\textwidth}{!}
{
\begin{tabular}{llll|ccl}
\toprule
\textbf{Methods} & \textbf{Narrator} & \textbf{Answerer} &  \textbf{Ego?} & 
\textbf{Subset} & \textbf{Full} & \textbf{Time (s)} \\ 
\midrule
MVU~\cite{mvu} & OWL-ViT~\cite{owl} & LLaVA-13B & \xmark & 60.3 & 37.6 & --\\
LangRepo~\cite{langrepo} & LaViLa~\cite{lavila} & Mixtral-8x7B & \cmark & 66.2 & 41.2 & -- \\
LLoVi~\cite{llovi} & LaViLa~\cite{lavila} & LLama3-8B & \cmark & 67.0 & 38.8 & --\\
\hline
Generative & SigLIP\&GPT-2 & LLama3-8B & \cmark & {66.5} & {37.4}& 48.2 \\
Ours & \our & LLama3-8B & \cmark & \textbf{70.4} & \textbf{43.1} & \textbf{2.3}\\
\bottomrule
\end{tabular}
}
\centering
\vspace{.5em}
\caption{\textbf{Video multiple-choice question-answering on Egoschema.} 
Each baseline consists of a narrator (to provide video information as reference) and an answerer (to respond to the question).
We report accuracy on both the subset and fullset, along with processing time per 180-second video.
}
\label{tab:egoschema}
\end{table}

\noindent\textbf{High-level Reasoning QA: EgoSchema.}
In Tab.\ref{tab:egoschema}, we present evaluation results on the Egoschema MCQ task. Our focus is not on surpassing state-of-the-art methods—most baselines \cite{videoagentmem, videoagentlong} rely on closed-source APIs (\eg, GPT-4o) or large-scale pretraining. Instead, we aim to validate the transferability of \our's reasoning capabilities.
To enable QA, we use \our~as a narrator to densely caption each long video, creating an informative document that LLMs can reference to answer questions accurately. An accurate narration should assist the LLM to correctly identify the answer.
We list highly relevant baselines, which has narrator and (open-source LLMs) answers.
We also develop generative baselines and compare their accuracy and runtime per 180 sec video. LLoVi~\cite{llovi}, which uses the same LLama3-8B models as the answerer and a GPT2 narrator~\cite{lavila} pretrained on ~\cite{ego4d}, being a comparable baseline.

Our \our~demonstrates an improvement margin (+3.4\% accuracy on the subset and +4.3\% on the full set), indicating more accurate narrations. Additionally, it achieves faster processing times per long video compared to generative baselines.

\subsection{Vocabularies Upgrading for OOV}
In Tab.~\ref{tab:oov}, we study the vocabulary upgrading strategy to address OOV problem. 
To design the experiments, we use HiREST's step captioning as benchmark, which includes several novel event (\eg, make a diet coke and mentos rocket) not seen in our vocabulary.
We include various baseline models for comprehensive comparison, selecting LLaVA-OV-0.5B~\cite{llavaov} -- the LMM in our agentic framework, conditioned on LLM's~\cite{qwen2.5} vocabulary.
For \our~models, we develop several variants with different vocabulary sizes and sources, including Ego4D, and {oracle HiREST vocabularies}, we compare produce narrations with groundtruth narrations by captioning metrics. 

\begin{table}[!h]
\small
    \centering
    \resizebox{0.48\textwidth}{!}{
    \begin{tabular}{llccc}
        \toprule
        Models & Vocab size & METEOR & CIDEr & SPICE \\ \midrule
        LLaVA-OV (Vid) & Qwen2 (152K)  & 1.2 & 1.0 & 0.1 \\
        LLaVA-OV (Img) & Qwen2 (152K)  & 2.3 & 4.2 & 2.5 \\
        \midrule
        VLog & Ego4D (0.8M) & 2.6&  5.4& 2.6\\
        VLog & COIN (778) & 3.0 & 6.9 & 2.3\\
        VLog & COIN +Upgrade (1223) & \textbf{4.2} & \textbf{12.6} & \textbf{3.0}\\
        \rowcolor[gray]{0.9}VLog & HiREST's task+Upgrade  & 4.8& 14.7 &3.2 \\
        \rowcolor[gray]{0.9}VLog & HiREST  & 5.8 & 21.2& 4.2\\
        \bottomrule
    \end{tabular}
    }
    \caption{\textbf{Key studies on Vocabulary upgrade strategies} for the HiREST Step Captioning Task across different vocabularies.}
\label{tab:oov}
\end{table}

In the experiment, we first observe that LLaVA-OV-0.5B does not achieve a high score with video inputs compared with image inputs, likely suffering hallucinations in small model size.
When come to \our, it is notable that the large Ego4D vocabulary underperforms compared to COIN, likely due to differences between egocentric and web-instruction videos. However, augmenting with 445 vocabulary items tailored for HiREST improves performance, validating this approach. Adding the oracle task name as a condition results in a slight performance boost, and using the oracle HiREST vocabulary achieves the best results. This demonstrates that vocabulary selection is crucial and that vocabulary upgrade strategy supports adaptation to unseen novel narrations.

\subsection{Ablation of Key Design Chocies}
\begin{figure}[!h]
    \centering
    \begin{minipage}{0.45\linewidth}
        \centering
        \includegraphics[width=\textwidth]{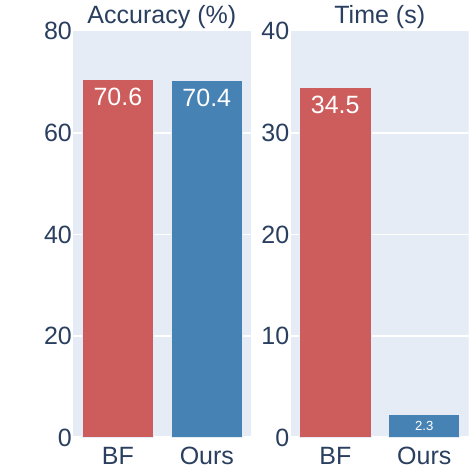}
        \caption{Ablation of Vocabulary Indexing Strategy}
        \label{fig:index:left}
    \end{minipage}\hfill
    \begin{minipage}{0.45\linewidth}
        \centering
        \includegraphics[width=\textwidth]{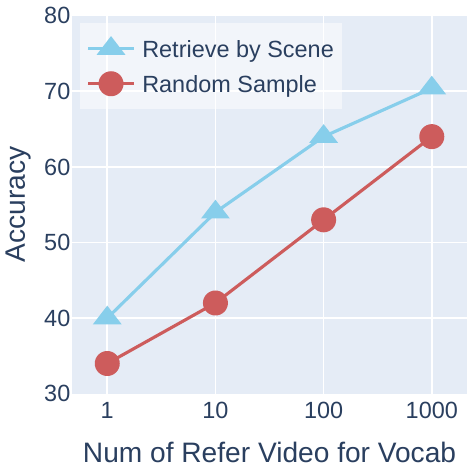}
        \caption{Ablation of Reference Video Number}
        \label{fig:index:right}
    \end{minipage}
    \caption{\textbf{Ablation studies of Vocabulary indexing on Egoschema QA.}
\textbf{Left:} our hierarchical indexing (by scene) is {15$\times$} faster than Brutal search over full vocabulary while keeping the accuracy.
\textbf{Right:} Using the same number of reference video for constructing narration vocabulary, selected by scene yields more relevant vocabularies.}
\end{figure}

\noindent\textbf{Hierarchical Indexing.}
In Fig.~\ref{fig:index:left}, we study the hierarchical indexing using the Egoschema val. set. ‘BF’ denotes Brutal search over the full vocabulary (0.8M), while ‘Hier.’ represents our proposed hierarchical indexing method.
Compared to a brutal search over a large space, our strategy is \textbf{15$\times$} faster, only averaging 2.3 seconds per video, and yields comparable accuracy. 

In Fig.~\ref{fig:index:right}, we compare \our~scene-based indexing with random indexing, adjusting the vocabulary size by varying the number of reference videos. We find that scene-based retrieval consistently yields a more reliable narration vocabulary and outperforms random sampling for the same vocabulary size. Accuracy improves with larger vocabulary sizes and additional reference videos, ensuring matched narrations appear more frequently.

\noindent\textbf{Definition of Retrieval Token.}
In Tab.\ref{tab:ablation:last}, we study how to define the memory token within \our, which could be the following variants:
(i) EOS, (ii) A learnable token, (iii) Pooling by all visual features.

\begin{table}[!h]
\small
    \centering
    \resizebox{0.4\textwidth}{!}{
    \begin{tabular}{lccc}
        \toprule
        Method &  Casual Retrieval & COIN-Step & COIN-Task\\ \midrule
        EOS & 4.3 & 51.3 & 93.1\\
        Learnable & 4.2& 50.9 & 92.5\\
        Pooling & \textbf{4.7} & \textbf{54.0} & \textbf{94.0}\\
        \bottomrule
    \end{tabular}
    }
    \caption{Ablation of different retrieval tokens initialization.}
    \vspace{-1em}
    \label{tab:ablation:last}
\end{table}

We found that the pooling strategy outperforms the EOS method, especially on COIN-step. However, for COIN-task, which requires longer input durations, the scores are closer. This suggests that for fine-grained action recognition, effective initialization is crucial.

\noindent\textbf{Can LMs knowledge help Generative Retrieval?}
In Tab.\ref{tab:knowledge}, we examine whether the pretrained knowledge of LLMs can enhance generative retrieval performance. For this, we develop several variants: GPT2 models with and without pretrained weights and models of different sizes to assess the strengths of LM's knowledge.

\begin{table}[!h]
\small
    \centering
    \resizebox{0.48\textwidth}{!}{
    \begin{tabular}{llccc}
        \toprule
        Method & Size &  Casual Retrieval & COIN-Step & COIN-Task\\ \midrule
        GPT2 & 124M & 4.7 & 54.0 & 94.0\\
        GPT2 (Random init.) & 124M & 3.8 & 51.5 & 91.0\\
        GPT2-Medium & 355M & {4.6} & {55.4} & \textbf{94.8} \\
        GPT2-Large & 774M & \textbf{4.9} & \textbf{56.4} & 93.2\\
        \bottomrule
    \end{tabular}
    }
    \caption{Ablation of LM's pretrained weights and sizes.}
    \vspace{-1em}    
    \label{tab:knowledge}
\end{table}

In general, pretrained and larger GPT2 models are beneficial, achieving higher scores on COIN-Step, which indicates that increased model size and richer textual knowledge enhance casual relationship for generative retrieval.

\subsection{Visualization}
Below, we provide an example to illustrate how \our~operates in a full decoding process. Given a video clip, \our~begins by identifying the scenario—‘doing yardwork’—then proceeds to the corresponding prefix vocabulary, resulting in “move a lawn mower machine” and finally completes the postfix as ‘on grass.’ Additional examples, illustrations on handling OOV cases and failure cases are available in the Supp. quantitative analysis Sec.
\begin{figure}[!h]
	\centering
	\includegraphics[width=1\linewidth]{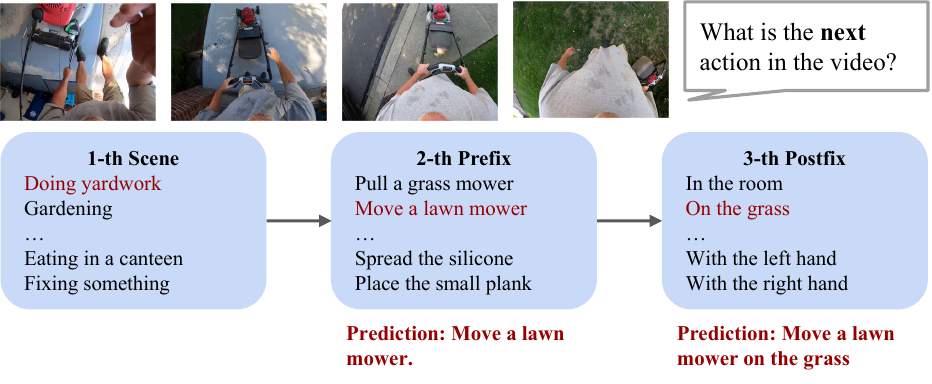}
\caption{{Illustration of \our's full decoding process.}
}
\label{fig:example:1}
\end{figure}
\section{Conclusion and Limitations}
We present \our, a novel framework for video streaming with a narration vocabulary. Built on a lightweight GPT-2 model, \our~introduces an innovative generative retrieval approach, combining causal modeling with retrieval efficiency. Additionally, \our~incorporates hierarchical vocabulary indexing and a vocabulary update mechanism. Experiments across several datasets demonstrate that \our~enables concise, contextually accurate, and efficient narrations, highlighting its potential for real-time video processing. 
We acknowledge that \our's design is constrained by its predefined vocabulary.
Future work will explore how to explore \our~to more diverse domains with a generalized query support.

\paragraph{Acknowledgements.}
This research is supported by the National Research Foundation, Singapore under its AI Singapore Programme (AISG Award No: AISG3-RP-2022-030).
\appendix
\section{Appendix}
\subsection{VidCab construction}
We begin by sourcing video clips from EgoClip~\cite{egovlp}, excluding any videos associated with downstream tasks such as Egoschema~\cite{egoschema}. Next, we clean the narrations by removing special tags like ‘\#C C’ and perform deduplication within each video, resulting in approximately 0.8M narrations. Using our Narration Pairing Encoding method, we generate a prefix set containing 0.6M entries and a postfix set with 5K entries, where the postfix is shared across all narrations and deduplicated.
Finally, we create a training and evaluation split at a 10:1 ratio, referred to as VidCab-Train and VidCab-Eval, respectively.

\subsection{Narration Pairing Encoding}
In the above algorithm, we display the process of our Narration Pairing Encoding algorithm, which mainly includes two parts:
\textbf{(i) Build Prefix Dictionary:} This step exhaustively enumerates all possible word combinations for each phrase to build a map between any prefix and the corresponding postfix narrations.
\textbf{(ii) Extract Prefixes and Postfixes:} For each narration, we determine whether other narrations share its full prefix. If not, we add it to the prefix list. If they do, we extract and collect the differing postfixes from the narrations that share its prefix.
\begin{figure}[!h]
	\centering
	\includegraphics[width=1\linewidth]{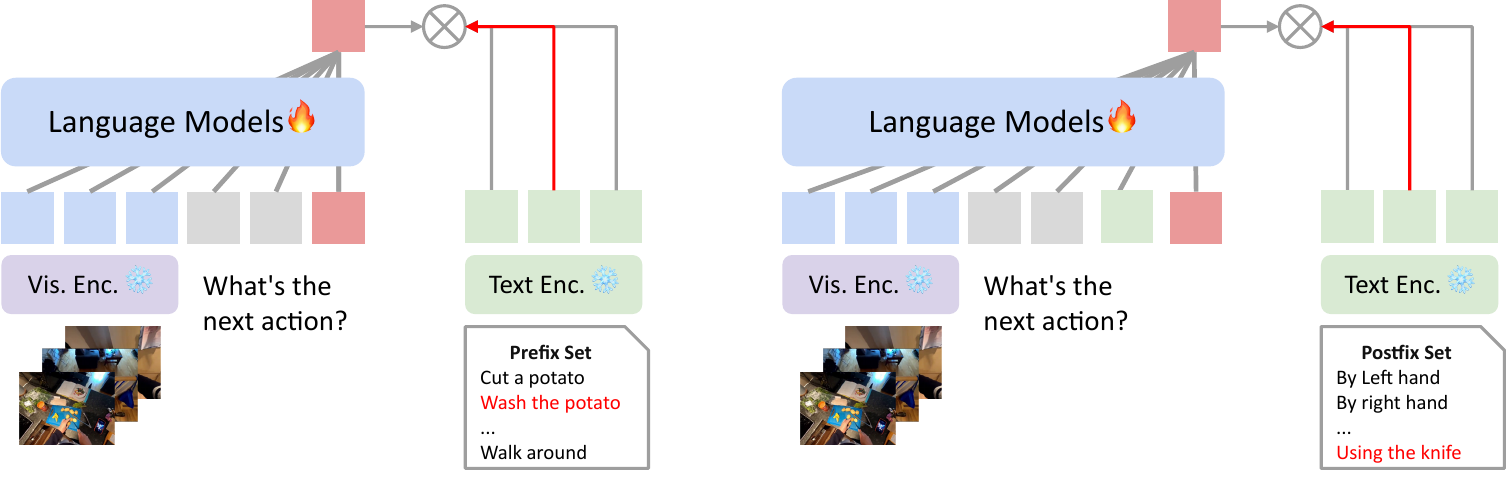}
\caption{{Illustration of \our's progressively decode prefix and postfix vocabulary respectively.}
}
\label{fig:decode}
\end{figure}

In Fig.~\ref{fig:decode}, we display how \our~progressively decode the prefix and postfix respectively. It first use the memory token to retrieve the prefix narration, and next it append the prefix narration and use the memory token to retrieve the postfix for a full narration.

\begin{algorithm}[h]
\caption{Narration Pairing Encoding}
\begin{algorithmic}[1]
\REQUIRE List of narrations $\mathcal{N}$
\ENSURE Prefix list $\mathcal{P}$, Postfix list $\mathcal{S}$

\STATE \textbf{Step 1: Build Prefix Dictionary}
\STATE Initialize prefix dictionary $\mathcal{D} \leftarrow$ empty
\FORALL{narration $n \in \mathcal{N}$}
    \STATE Split $n$ into words $[w_1, w_2, \dots, w_k]$
    \FOR{$i = 1$ to $k$}
        \STATE Prefix $p \leftarrow w_1\, w_2\, \dots\, w_i$
        \STATE Add $n$ to $\mathcal{D}[p]$
    \ENDFOR
\ENDFOR

\STATE \textbf{Step 2: Extract Prefixes and Postfixes}
\STATE Initialize prefix list $\mathcal{P} \leftarrow \emptyset$, postfix list $\mathcal{S} \leftarrow \emptyset$
\FORALL{narration $n \in \mathcal{N}$}
    \IF{$\mathcal{D}[n]$ contains only $n$}
        \STATE Add $n$ to $\mathcal{P}$
    \ELSE
        \FORALL{other narration $s \in \mathcal{D}[n]$}
            \IF{$s \neq n$}
                \STATE Get suffix $t \leftarrow$ Remove prefix $n$ from $s$
                \STATE Add $t$ to $\mathcal{S}$
            \ENDIF
        \ENDFOR
    \ENDIF
\ENDFOR

\end{algorithmic}
\end{algorithm}

\subsection{Vocabulary Update Templates}
Below, we attach the prompts for Qwen2.5~\cite{qwen2.5}, which is used for produce narrations directly.

\begin{tcolorbox}[boxrule=0pt, colframe=white, sharp corners, left=1mm, right=1mm, top=0.2mm, bottom=0.2mm]
\footnotesize
{
List possible short actions that could take place in the scene.
Write each action as a short narration (a verb with a noun). Separate by ';'
The following is examples.
\\
\textbf{scene:} In the heart of the kitchen, a man skillfully slices into a ripe mango, its golden flesh gleaming under the light.
\\
\textbf{narration:} Slice mango; Hold knife; Cut mango; Place seed; Wipe counter; Drop pieces; Grip mango; Rest knife; Smell mango; Gather chunks.
\\
\\
\textbf{scene:} A woman sits by the fireplace, knitting a scarf as the flames crackle warmly in the background.
\\
\textbf{narration:} Knit scarf; Hold needles; Loop yarn; Adjust thread; Pull stitch; Rest hands; Drop yarn; Smell smoke; Listen flames; Rub hands; Fold scarf; Gather wool; Stare fire; Sit still; Tap needle.
\\
\\
\textbf{scene:} {\texttt{\{scene\}}}
\\
\textbf{narration:}
}
\end{tcolorbox}
The {\texttt{\{scene\}}} is output by LLaVA-OV-0.5B~\cite{llavaov} with prompt: “What is the overall activity in the scene? Answer briefly in one sentence.”

\subsection{Experimental Settings}
Our SigLIP model~\cite{siglip} is based on \texttt{google/siglip-so400m-patch14-384}. During training, we fully fine-tune the GPT-2 model with a batch size of 32, a learning rate of 3e-4, and a sampling rate of 8 frames per short video clip. For long videos, such as those in the EgoSchema dataset~\cite{egoschema}, we do not compress the entire video into a single embedding. Instead, we uniformly sample long videos into multiple fixed-length clips (1 second each) and process them in a streaming fashion.

\subsection{Complexities Analysis}
Let us clarify each term when generating $N$ narrations:
\textbf{(i) Encoding:} We embed the entire vocabulary once and then reuse it -- $O(1)$.
\textbf{(ii) Decoding:} This should be $O(\alpha N)$, where $\alpha$ is the speed per decoding step.
\textbf{(iii) Upgrading} (optionally): $O(C)$, where $C$ is the upgrade times ($C\ll N$).
For a large $N$, the overall complexity $O(1)+O(\alpha N)+O(C)\rightarrow O(\alpha N)$ \textit{remains efficient as the encoding and upgrading costs become negligible.}
Below is the timing analysis on 4.6K VidCap-Eval:

\begin{table}[!htp]
\centering
\resizebox{\columnwidth}{!}
{
\begin{tabular}{ll|ccc|cc}
Models  &  Vocab. size & \textbf{Encoding}(s) & \textbf{Decoding}(s) & \textbf{Upgrading}(s) & \textbf{R@1} & \textbf{Total (s)} \\
\hline
GPT-2 & 32K & -- & 207.8 & -- & 7.9 & 207.8 \\
VLog & 4.6K & \textcolor{gray}{3.6} & 10.4 & -- & 12.4 & \textbf{14.0} \\
VLog & 4.6K (+486) & \textcolor{gray}{3.6} & 10.5 & 38.4 & \textbf{13.7} & 52.4 \\
\end{tabular}
}
\end{table}

\subsection{Subwords \textit{v.s.} Narration Vocabulary on Easy \textit{v.s.} Complex tasks?}
Our VLog is {prioritize task-specific efficiency over generalist models}. We compare the two in the below Table.
\begin{table}[!htp]
\centering
\resizebox{\columnwidth}{!}
{
\begin{tabular}{l|lllll}
  & Domain  & Vocabulary & Backbone  & Decoding & Highlights \\
\hline
VideoLLMs & General & Subwords & LLMs (2B+)  & Token Gen. & Multi-Purpose \\
\textbf{VLog} & \textbf{Specific} & \textbf{Narrations} & GPT-2 (\textbf{345M}) & \textbf{Retrieval} & \textbf{Efficiency} \\
\end{tabular}
}
\end{table}

Whether `Subwords-' or `Narration-' Vocabulary is depends on {how tasks define minimal semantic units for videos}. 
Subwords capture every detail but may be redundant for long videos, while narrations offer event contexts quickly but may miss finer details.
{To balance  expressive granularity and efficiency}, an idea is to {{cooperate two fashions}} like {our vocab. upgrading} or {retrieve narration first and then generate minimal subwords as needed.}
We are interested in further exploring the latter.

\subsection{Improvment by Stronger LLM}
We chose GPT-2 because its \textit{simplicity and lightweight nature} make it a representative baseline.
To demonstrate scaling, we upgraded GPT-2 to Qwen2-7B, resulting in significant performance gains, and beat its comparable baseline Qwen2-VL-7B.

\begin{table}[!htp]
\centering
\scriptsize
{
\begin{tabular}{l|cccc}
Models & Size & EgoSchema QA val. & Decoding Time (s) \\
\hline
Qwen2-VL & 7B & 72.8 & 79.4\\
VLog (GPT-2) & 345M & 70.4 & \textbf{2.3} \\
VLog (Qwen2) & 7B & \textbf{74.8} & 6.4 \\
\end{tabular}
}
\end{table}

\section{Qualitative Examples}
\subsection{VLog for Reasoning Retrieval}
\begin{figure*}[!t]
	\centering
	\includegraphics[width=1\linewidth]{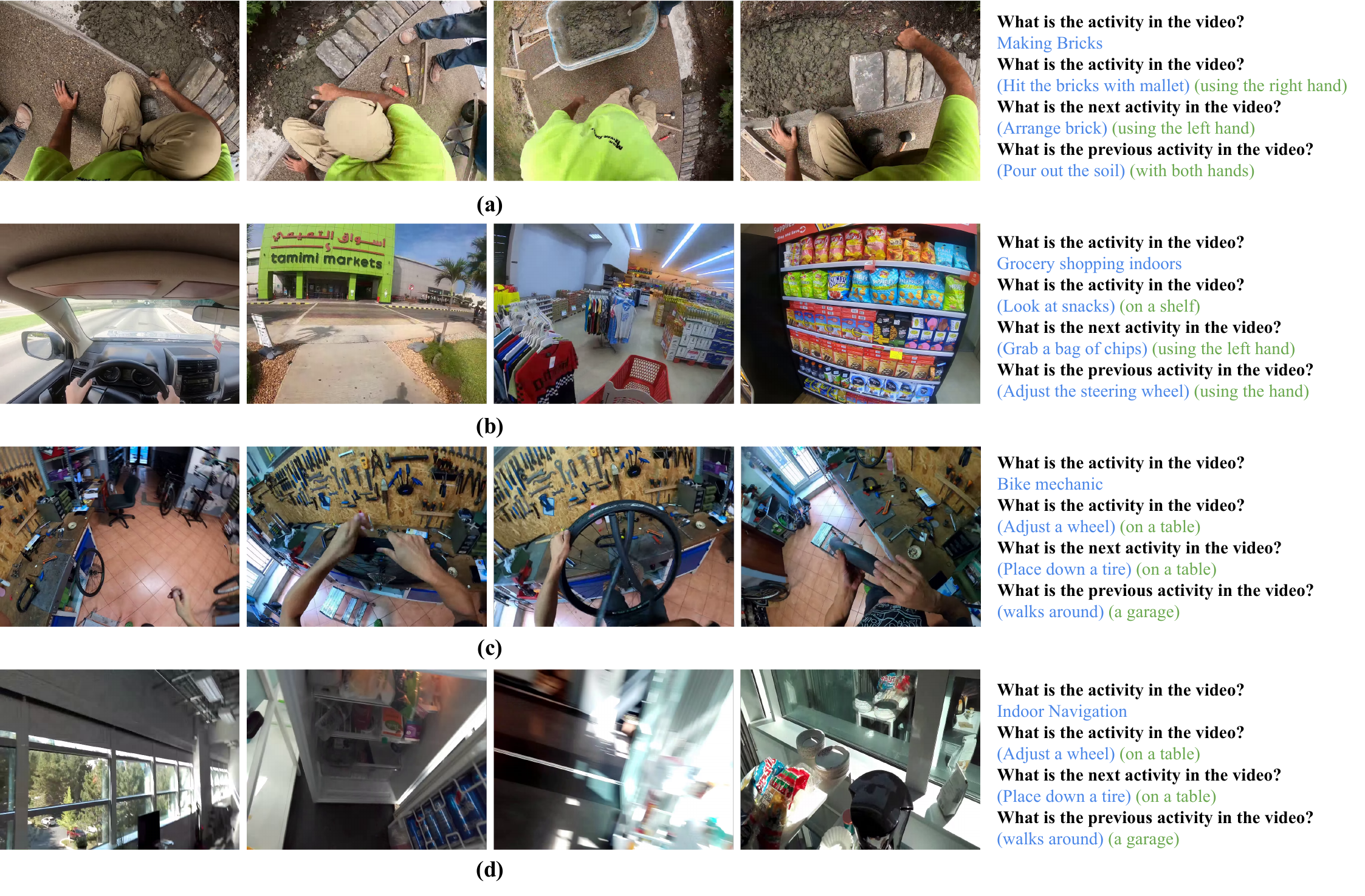}
\caption{\textbf{\our~enables retrieval through reasoning}, conditioned on different queries. Blue represents prefixes, while green represents postfixes.}
\label{fig:example:a}
\end{figure*}

In Fig.\ref{fig:example:a}, we illustrate how \our~retrieves the vocabulary (blue indicating prefixes and green indicating postfixes) conditioned on different queries. For instance, in example (b), the query “What is the next activity in the video?” retrieves “Grab a bag of chips using the left hand” as a result, while the query “What is the previous activity in the video?” retrieves “Adjust the steering wheel using the hand” as a result, demonstrating \our~'s capability to infer relationships between sequential events.

\subsection{How does Vocabulary Updating work?}
\begin{figure*}[!t]
	\centering
	\includegraphics[width=1\linewidth]{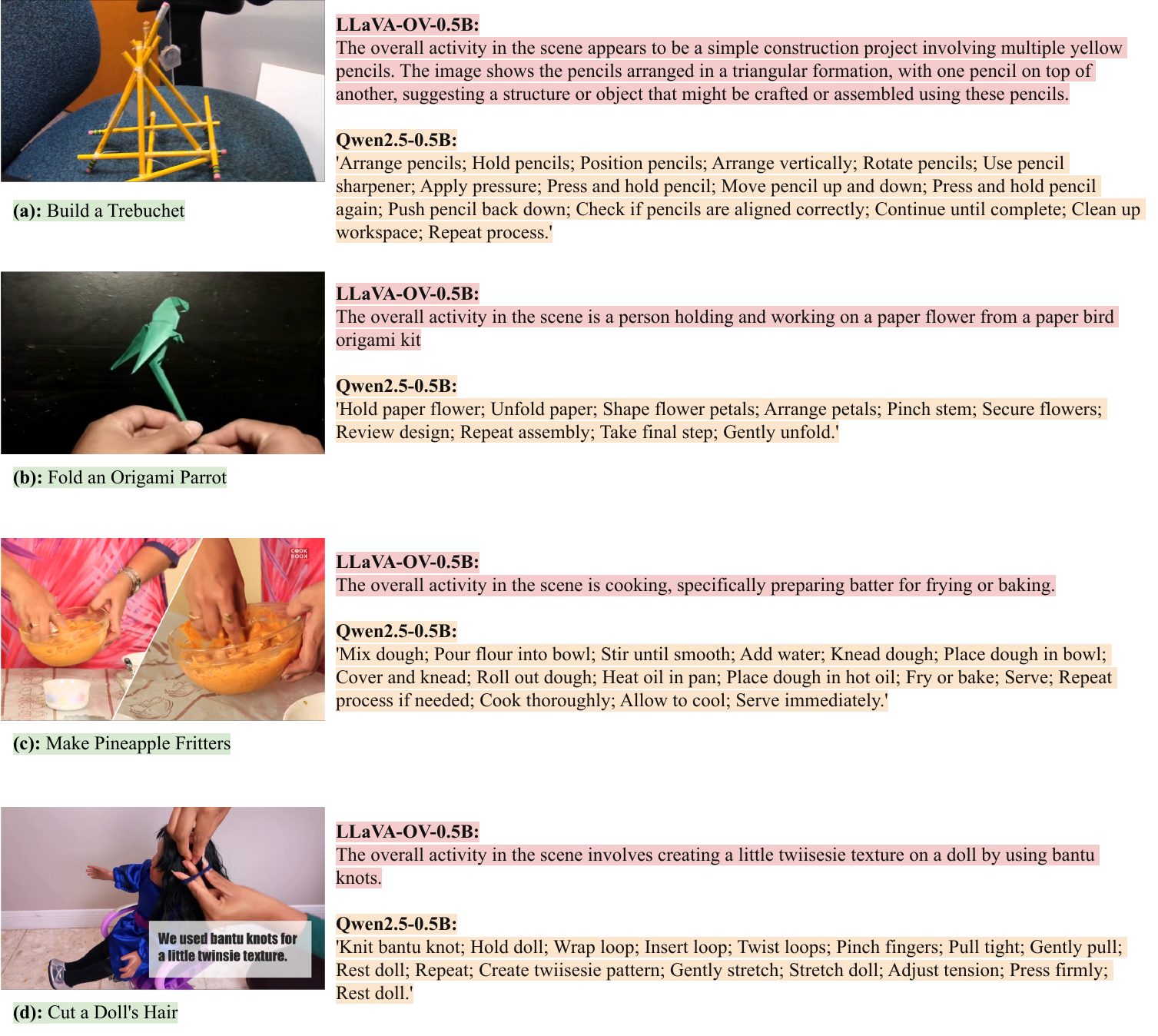}
\caption{\textbf{Illustration of \our's vocabulary updating process.} Given the first frame of a video clip, LLaVA-OV~\cite{llavaov} provides a brief initial description, which is then passed to Qwen2.5~\cite{qwen2.5} to generate and expand the possible vocabulary.}
\label{fig:example:b}
\end{figure*}

In Fig.\ref{fig:example:b}, we demonstrate how \our's vocabulary updating process effectively expands its descriptive range. Given the first frame of a video clip, LLaVA-OV\cite{llavaov} generates an initial brief description, which is then passed to Qwen2.5~\cite{qwen2.5} to imagine and expand possible vocabulary terms. For instance, in (a), LLaVA-OV identifies a simple construction project involving multiple yellow pencils, and Qwen2.5 extends this by generating potential actions such as “Arrange pencils" and “Hold pencils," which collectively capture most events in the video.

However, limitations still exist with the models. For example, in (c), while the activity “Make Pineapple Fritters" is identified, the model struggles to detect the specific ingredient “pineapple," making it challenging for the expanded vocabulary to recognize or describe the desired object accurately. These challenges highlight areas for improvement in object-specific vocabulary generation.

\subsection{Limtation by VLog.}
\begin{figure}[!h]
	\centering
	\includegraphics[width=1\linewidth]{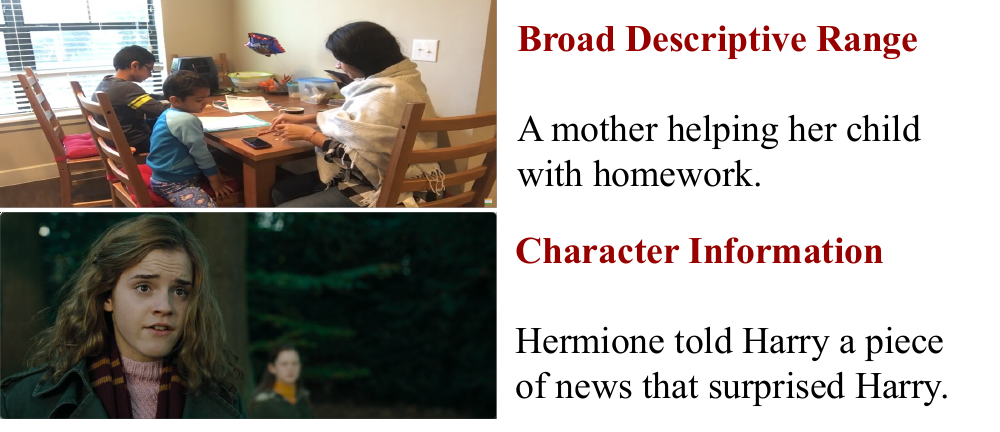}
\caption{\our~still fail to capture the video with broad descriptive range or high-level information \eg~characters.
}
\label{fig:falture}
\end{figure}
We acknowledge that \our~still has limitations, as illustrated in \ref{fig:falture}. For example, when videos have a broad expressive range, such as those involving multiple individuals or focusing on different aspects depending on subjective interpretation, it becomes challenging to rely on a narration-wise closed-set vocabulary. Additionally, in more complex scenarios, such as movies, where character information and dialogues play a central role, the current approach struggles. These cases may require a return to a generalist model capable of handling subword tokens for richer representations.

{\small
\bibliographystyle{ieee_fullname}
\bibliography{main}
}

\end{document}